\pdfoutput=1

\documentclass[preprint,11pt,times]{elsarticle}

\usepackage{amsmath,amssymb}
\usepackage{graphicx}
\usepackage{booktabs}
\usepackage{multirow}
\usepackage[hidelinks]{hyperref}
\usepackage{xcolor}
\usepackage{tikz}
\usetikzlibrary{arrows.meta,positioning,fit,calc}
\usepackage{algorithm}
\usepackage{algpseudocode}
\makeatletter
\def\ps@pprintTitle{\let\@oddhead\@empty \let\@evenhead\@empty
  \def\@oddfoot{\centerline{\thepage}} \let\@evenfoot\@oddfoot}
\makeatother

\newcommand{\shrinkbox}[1]{\resizebox{\ifdim\width>\linewidth\linewidth\else\width\fi}{!}{#1}}
\DeclareFontFamily{T1}{cmtt}{}
\DeclareFontShape{T1}{cmtt}{m}{n}{<->ssub*txtt/m/n}{}
\usepackage{array}
\usepackage{float}

\begin{document}

\begin{frontmatter}

\title{Deep Reinforcement Learning on Item-Compatibility Graphs\\for One-Dimensional Bin Packing}

\author[bau]{M.\ Asl\i{} Ayd\i n\corref{cor}}
\ead{asli.aydin@bau.edu.tr}
\cortext[cor]{Corresponding author.}
\affiliation[bau]{organization={Department of Management Engineering, Bah\c{c}e\c{s}ehir University},
  city={Istanbul}, country={T\"urkiye}}

\begin{abstract}
The one-dimensional bin packing problem (1D-BPP) is a classical NP-hard combinatorial optimization problem with applications ranging from logistics and manufacturing to cloud resource management. Although deep reinforcement learning (DRL) has become a competitive paradigm for data-driven optimization, most learned packing methods target 2D and 3D variants, and intelligent learned solvers for 1D-BPP remain scarce. In this paper, we present a novel end-to-end, size-agnostic graph reinforcement learning framework for 1D-BPP. We formulate the packing process as a Markov decision process on an item-compatibility graph, serving as a structural knowledge representation in which every action merges two partial bins that fit together. A graph neural network actor--critic policy extracts relational features from this representation and is trained through reinforcement learning and decoded by stochastic beam search, enabling a single trained model to generalize zero-shot to instances of any size. We conduct a systematic empirical study across graph encoders, DRL algorithms, reward functions, training distributions, and hyperparameters. Evaluated zero-shot on the full BPPLIB benchmark against a constructive heuristic, a grouping genetic algorithm, and recent learned methods, our data-driven policy lowers the mean optimality gap of the constructive heuristic from 2.66\% to 2.31\%, with the largest gains on structured instances. Against learned baselines evaluated on the same benchmark, it attains a lower gap on most of the nine families and is far more stable across instance distributions. On the hardest benchmark family, it outperforms a state-of-the-art learned solver that relies on column generation and integer programming, while using no solver at all. A grouping genetic algorithm remains ahead overall, and we analyze where and why the residual gap arises.
\end{abstract}

\begin{keyword}
Bin packing \sep Deep reinforcement learning \sep Graph neural networks \sep Combinatorial optimization \sep Item-compatibility graph 
\end{keyword}

\end{frontmatter}

\section{Introduction}
\label{sec:intro}

The one-dimensional bin packing problem (1D-BPP) asks for the assignment of $n$ items with known weights $w_1,\dots,w_n$ to the smallest possible number of identical bins of capacity $C$ such that the total weight in every bin does not exceed $C$~\cite{martello1990knapsack}. Despite its simple statement, 1D-BPP is the core formulation behind a wide range of industrial decisions. In freight transportation and supply-chain management it models the cargo loading onto the fewest possible trucks or containers~\cite{kbs_transport_survey2021,martello1990knapsack}. In manufacturing it coincides with the unit-demand one-dimensional cutting stock problem (CSP), which governs the cutting of steel bars, paper rolls or glass tubes with minimum trim loss~\cite{delorme2016bin,wascher2007}. In information technology it underlies virtual machine placement and server consolidation in data centres~\cite{bansal2016,bein2011}. The problem also continues to attract methodological work in operations research, including recent exact approaches to variants that couple packing with scheduling objectives~\cite{marinelli2025bpvpt}.

1D-BPP is NP-hard~\cite{gareyjohnson1979}. Exact methods such as branch-and-price, arc-flow formulations and the BISON procedure~\cite{scholl1997bison,valerio1999,delorme2016bin} solve moderate instances to proven optimality but struggle on large or structurally hard instances~\cite{delorme2018bpplib}. In this case, practitioners often use fast constructive heuristics, most notably First-Fit Decreasing (FFD) and Best-Fit Decreasing (BFD). The idea is to sort the items in non-increasing order first and then place each item into the first or the tightest feasible bin~\cite{johnson1974,martello1990knapsack,coffman2013,fleszar2002}. Their worst-case guarantees are well established~\cite{dosa2007} and on many benchmark instances FFD is within a fraction of a bin of the optimum. Where higher quality is required, metaheuristics ranging from local search and simulated annealing to grouping genetic algorithms search the combinatorial space and close most of the remaining gap~\cite{falkenauer1996hybrid,kucukyilmaz2018,munien2020}. However, metaheuristics are comparatively expensive. They depend on hand-crafted problem-specific operators, and solve every new instance from scratch without retaining any transferable knowledge.

Over the past decade, deep reinforcement learning (DRL) and neural combinatorial optimization have emerged as a data-driven alternative ~\cite{bengio2021,mazyavkina2021reinforcement,vesselinova2020}. Pointer networks, graph neural networks (GNNs) and attention models achieve competitive results on routing problems such as the travelling salesman and vehicle routing problems~\cite{vinyals2015pointer,bello2016neural,khalil2017learning,kool2019attention,nazari2018}. In packing, learned methods have focused almost exclusively on two- and three-dimensional variants~\cite{li2022rcql,zhao2024dmrl,graphpack2023}, where spatial layouts offer natural inductive biases and classical heuristics leave substantial room for improvement.

End-to-end learning for 1D-BPP, in contrast, has received little attention~\cite{ml4bpp2023,zhang2023review,rlbpp_review2025}. There are two main reasons for this. First, a 1D instance is just a set of numbers with no coordinates or adjacency. So sequence models must cope with permutation invariance and dynamic capacity constraints without any spatial prior. Second, because FFD is already near-optimal on average, a learned policy risks collapsing to a sorting rule unless it discovers non-myopic item combinations. Hence, some existing methods train a network to predict only an item \emph{ordering} and leave placement to a heuristic~\cite{fang2023,huang2026conveyor}. Some other method embeds a bipartite GNN inside a column-generation loop that solves linear and integer programs at every inference step~\cite{shi2025combination}.

This paper asks whether a pure, end-to-end graph DRL solver can learn high-quality packing policies for 1D-BPP. We propose a framework that first turns a 1D-BPP instance into an \emph{item-compatibility graph}~\cite{sensarma2017} whose nodes are (partial) bins and whose edges join pairs whose combined load fits into one bin. Then packing becomes a sequential edge-merge Markov decision process (MDP) where each action contracts one edge, and the episode ends when no compatible pair remains. Every node is described only by ratios, its relative load and its normalized degree. The network scores each candidate merge from the two nodes it joins and summarizes the whole graph by averaging over its nodes. Hence no part of the model depends on how many items an instance contains. A single trained policy therefore transfers zero-shot to instance sizes and families never seen during training.

Beyond its algorithmic performance, the proposed end-to-end framework connects classical combinatorial optimization with modern intelligent decision support systems. Casting 1D-BPP onto an item-compatibility graph provides a structural knowledge representation of the combinatorial space. On this representation the graph neural network extracts relational knowledge about feasible item combinations and learns a data-driven optimization policy without hand-crafted heuristic rules. The trained size-agnostic model can therefore serve as an adaptable decision component within intelligent decision support systems in which packing decisions recur at scale and must be made without a solver in the loop.

The contributions of this paper are as follows.
\begin{itemize}
    \item We define 1D-BPP as an MDP on an item-compatibility graph with a size-invariant state representation whose return is objective-equivalent to the bin count.
    \item We develop an end-to-end, size-agnostic graph reinforcement learning framework. By utilizing relative capacity and topological degree features, the model achieves strong zero-shot generalization to unseen problem scales.
    \item We compare three graph encoders (GCN, GAT, GIN) and six RL algorithms (PPO, A2C, REINFORCE, DQN, SAC, SARSA) selecting all design choices on in-distribution validation data rather than on the benchmark.
    \item We evaluate a single uniform-trained model on all complete BPPLIB families against FFD, a grouping genetic algorithm and recent learned solvers on the identical benchmark and metric. We analyse where the learned policy helps, where it does not, and why.
\end{itemize}

The remainder of the paper is organized as follows. Section~\ref{sec:related} reviews classical, metaheuristic and learning-based approaches to 1D-BPP. Section~\ref{sec:problem} states the problem formally and introduces the item-compatibility graph. Section~\ref{sec:method} presents the MDP, the DRL architecture, and the training and decoding procedures. Section~\ref{sec:experiments} describes the computational experiments and reports the results. Section~\ref{sec:discussion} discusses their implications and Section~\ref{sec:conclusion} concludes with limitations and future work.

\section{Related work}
\label{sec:related}

\subsection{Exact and heuristic methods}
The study of bin packing and cutting stock problems originates with Kantorovich's production-planning formulation~\cite{kantorovich1960} and the delayed column generation of Gilmore and Gomory~\cite{gilmore1961}. Dyckhoff~\cite{dyckhoff1990} and W\"ascher et al.~\cite{wascher2007} established the standard typologies of cutting and packing problems, and Martello and Toth~\cite{martello1990knapsack} provided the reference integer programming models, lower bounds and branch-and-bound algorithms. Modern exact solvers include the arc-flow formulation of Val\'erio de Carvalho~\cite{valerio1999,carvalho2002} and the BISON procedure of Scholl et al.~\cite{scholl1997bison}. Delorme et al.~\cite{delorme2016bin,delorme2018bpplib} surveyed mathematical models and instances in the BPPLIB library, which we use as the test bed. Exact methods solve moderate instances efficiently but their effort grows sharply on instances with tight capacity structure. For real-time use, constructive heuristics remain the practical standard. FFD and BFD~\cite{johnson1974} admit tight worst-case bounds~\cite{dosa2007,coffman2013}, and have been refined by later constructive schemes~\cite{fleszar2002}.

\subsection{Grouping metaheuristics}
When constructive heuristics leave an unacceptable residual gap and exact solvers are too slow, metaheuristics search the solution space directly. Early studies applied simulated annealing and tabu search~\cite{munien2020}. But item-level encodings suffer from the fact that moving single items between bins frequently violates capacity. Falkenauer's Grouping Genetic Algorithm (GGA) and its hybrid variant HGGA~\cite{falkenauer1996hybrid} instead manipulate whole bins with group-level crossover and mutation, and remain the standard high-quality reference for 1D-BPP. Later work developed parallel and cooperative grouping metaheuristics~\cite{kucukyilmaz2018} and hybrid evolutionary schemes~\cite{munien2020,gonzalez2023}. However, metaheuristics offer no quality or convergence guarantees and are comparatively expensive. They are also memoryless, meaning that nothing learned on one instance transfers to the next.

\subsection{Learning-based methods for bin packing}
These limitations, together with the success of DRL on hard sequential decision problems~\cite{silver2016}, have motivated a line of work often called neural combinatorial optimization. Instead of relying on hand-designed rules, it learns solution policies for combinatorial problems directly from data~\cite{bengio2021,mazyavkina2021reinforcement}. Pointer networks~\cite{vinyals2015pointer,hu2017solving}, policy-gradient training~\cite{bello2016neural}, structure-to-vector GNNs~\cite{khalil2017learning} and attention models~\cite{kool2019attention,nazari2018} produce high-quality solutions on routing problems, which are naturally defined on graphs. In bin packing, learned solvers have concentrated on 2D and 3D variants and on strip packing~\cite{li2022rcql,zhao2024dmrl,graphpack2023,ml4bpp2023}, where spatial layouts create geometric sub-problems that heuristics handle poorly. Recent surveys of learning-based methods for bin packing make the same observation~\cite{ml4bpp2023,rlbpp_review2025,zhang2023review}. Almost all of the effort has gone into the multi-dimensional variants, and the one-dimensional problem has attracted very little of it. The few 1D approaches are not end-to-end constructors. Instead, they learn an item ordering for a heuristic decoder~\cite{fang2023,huang2026conveyor}, or combine a bipartite graph convolutional network with column generation and integer programming so that linear and integer programs are solved at every inference step~\cite{shi2025combination}. 

\subsection{Graph representations of 1D-BPP}
Applying a GNN policy to 1D-BPP requires projecting weights and capacities onto an explicit graph. The arc-flow network~\cite{valerio1999} is one option, but its nodes index integer capacity levels, so its size scales with $C$ and its action semantics are awkward for a stationary sequential decision process. The item-compatibility graph~\cite{sensarma2017} instead places one node per item (or partial bin) and joins two nodes whenever their combined load fits into a bin. Its edge set is exactly the set of feasible pairwise merges, and a merge updates the graph locally, which makes it a natural state space for reinforcement learning.

Our work differs from prior learned 1D-BPP methods in two respects. First, the MDP is defined directly on the item-compatibility graph with a size-independent state, so one trained policy applies to instances of any size. Second, every packing decision is produced by the learned policy rather than by a heuristic placement rule~\cite{fang2023} or by a mathematical programming loop~\cite{shi2025combination}. We evaluate this design on the complete BPPLIB benchmark to establish where learned constructive policies succeed and where their limits lie relative to constructive heuristics, grouping metaheuristics and existing learned solvers.

\section{Problem definition}
\label{sec:problem}

\subsection{One-dimensional bin packing problem}
An instance of 1D-BPP is a pair $(W,C)$, where $W=\{w_1,\dots,w_n\}$ is a set of $n$ items with integer weights $0<w_i\le C$ and $C$ is the common capacity of an unlimited supply of identical bins. A packing assigns every item to exactly one bin such that the total weight in every bin is at most $C$ and the objective is to minimize the number of bins used. With binary variables $y_k$ (bin $k$ is used) and $x_{ik}$ (item $i$ is placed in bin $k$), for $k\in\{1,\dots,n\}$, the integer programming formulation of 1D-BPP is~\cite{martello1990knapsack}
\begin{align}
\min\ & \sum_{k=1}^{n} y_k \label{eq:ilp_obj}\\
\text{s.t.}\ & \sum_{k=1}^{n} x_{ik} = 1, && i=1,\dots,n, \label{eq:ilp_assign}\\
& \sum_{i=1}^{n} w_i x_{ik} \le C\,y_k, && k=1,\dots,n, \label{eq:ilp_cap}\\
& x_{ik},\,y_k\in\{0,1\}, && i,k=1,\dots,n. \label{eq:ilp_bin}
\end{align}
The continuous relaxation of~\eqref{eq:ilp_obj}--\eqref{eq:ilp_bin} yields the trivial lower bound $L_1=\lceil \sum_i w_i / C\rceil$.
\subsection{Item-compatibility graph}
\label{sec:problem:graph}
An instance $(W,C)$ has no graph structure of its own. Following~\cite{sensarma2017}, we build the \emph{item-compatibility graph} of the instance as an undirected graph $G=(V,E)$ where
\begin{equation}
V=\{1,\dots,n\},\qquad E=\{(i,j)\,:\,i<j,\ w_i+w_j\le C\}.
\label{eq:compat}
\end{equation}
An edge between two vertices indicates that the two items may share a bin. A packing corresponds to a partition of $V$ into vertex sets $S_1,\dots,S_b$ such that every $S_k$ is a clique of $G$ and $\sum_{i\in S_k} w_i\le C$. Minimizing the number of bins used therefore reduces to a capacity-constrained minimum clique partition of $G$.

\begin{figure}[htbp]
\centering
\scalebox{0.7}{
\begin{tikzpicture}[
    node distance=2cm,
    itemnode/.style={circle, fill=orange!40, draw=none, minimum size=8mm, align=center, font=\small},
    edgelabel/.style={fill=white, inner sep=1pt, font=\scriptsize}
]
\begin{scope}[xshift=0cm]
    \node[itemnode, label=below:{$w_1=1$}] (d1) at (1.5, 0) {$d_1$};
    \node[itemnode, label=above:{$w_2=2$}] (d2) at (1.5, 2.5) {$d_2$};
    \node[itemnode, label=above:{$w_3=4$}] (d3) at (3.5, 2.5) {$d_3$};
    \node[itemnode, label=below:{$w_4=5$}] (d4) at (3.5, 0) {$d_4$};
    \node[itemnode, label=below:{$w_5=9$}] (d5) at (0, 1.25) {$d_5$};
    \node at (2.25, -1.3) {(a)};
\end{scope}
\begin{scope}[xshift=6cm]
    \node[itemnode, label=below:{$w_1=1$}] (d1) at (1.5, 0) {$d_1$};
    \node[itemnode, label=above:{$w_2=2$}] (d2) at (1.5, 2.5) {$d_2$};
    \node[itemnode, label=above:{$w_3=4$}] (d3) at (3.5, 2.5) {$d_3$};
    \node[itemnode, label=below:{$w_4=5$}] (d4) at (3.5, 0) {$d_4$};
    \node[itemnode, label=below:{$w_5=9$}] (d5) at (0, 1.25) {$d_5$};
    \draw[thick] (d2) -- (d5) node[midway, edgelabel] {$k_1$};
    \draw[thick] (d1) -- (d5) node[midway, edgelabel] {$k_2$};
    \draw[thick] (d1) -- (d2) node[midway, edgelabel] {$k_3$};
    \draw[thick] (d2) -- (d4) node[pos=0.3, edgelabel] {$k_4$};
    \draw[thick] (d2) -- (d3) node[midway, edgelabel] {$k_5$};
    \draw[thick] (d1) -- (d3) node[pos=0.3, edgelabel] {$k_6$};
    \draw[thick] (d1) -- (d4) node[midway, edgelabel] {$k_7$};
    \draw[thick] (d3) -- (d4) node[midway, edgelabel] {$k_8$};
    \node at (2.25, -1.3) {(b)};
\end{scope}
\end{tikzpicture}
}
\caption{Item-compatibility graph of a five-item instance with $C=11$. (a) Items as isolated nodes. (b) Edges join pairs whose combined weight does not exceed $C$.}
\label{fig:example_graph}
\end{figure}
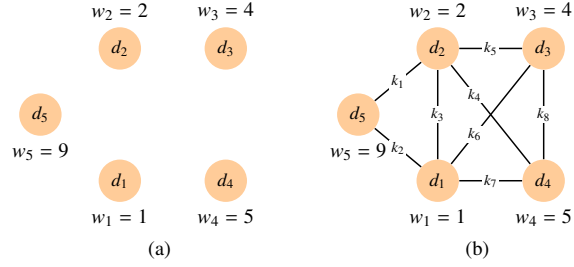

\paragraph{Illustrative example} Figure~\ref{fig:example_graph} shows how the item-compatibility graph is constructed for a small sample instance with $n=5$ items, $W=\{1,2,4,5,9\}$ and $C=11$. The eight edges $k_1,\dots,k_8$ of the resulting graph denote the pairs of items whose weights sum to at most 11.

\section{Methods}
\label{sec:method}

\subsection{MDP formulation}
\label{sec:method:mdp}
We model packing as a deterministic, episodic MDP $(\mathcal{S},\mathcal{A},T,R)$ in which every step contracts one edge of the current graph.

\paragraph{State} The state at step $t$ is the graph $G_t=(V_t,E_t)$ whose nodes are the current partial bins. Node $i$ carries a load $\ell_i$, initialized to $w_i$, and two nodes are adjacent while their combined load fits into a bin, $(i,j)\in E_t \iff \ell_i+\ell_j\le C$. Hence $G_0$ is the item-compatibility graph~\eqref{eq:compat}. Each node is described by the two-dimensional feature vector
\begin{equation}
x_i=\Big[\ \frac{\ell_i}{C},\ \ \frac{\deg_t(i)}{\max_{k\in V_t}\deg_t(k)}\ \Big],
\label{eq:features}
\end{equation}
i.e.\ its relative load and its degree normalized by the largest degree in $G_t$. Both entries are ratios in $[0,1]$, which makes the representation independent of $n$ and of the absolute scale of $C$. The normalized degree measures how many merge options a node still has. A node with a high value fits with almost everything and can wait, while a node with a low value has only a few compatible partners left and risks ending up as a bin on its own.

\paragraph{Action} The action set is the current edge set, $\mathcal{A}(s_t)=E_t$. Choosing $a_t=(i,j)$ merges the two endpoints into a single node of load $\ell_i+\ell_j$, i.e.\ it packs the two partial bins together. Feasibility is guaranteed by construction, so no masking or penalty for infeasible actions is required.

\paragraph{Transition} Transitions are deterministic. The merged node replaces its two endpoints. Its incident edges are recomputed from the new load, so that any edge that would now violate the capacity is removed. Since exactly one node disappears per step, an episode has at most $n-1$ steps and the number of bins at termination equals $n$ minus the number of merges.

\paragraph{Reward} We use a per-step reward $R(s_t,a_t)=+1$ for every merge and $0$ at termination, so the return of an episode is the number of merges. Since the number of bins used equals $n$ minus the total number of merges, maximizing the cumulative reward (the total number of merges) is equivalent to minimizing the number of bins used. We define and compare two alternative reward designs in Section~\ref{sec:exp:ablation:reward}. 

\paragraph{Terminal state} An episode ends when $E_t=\varnothing$, i.e.\ no two remaining bins can be combined. At that point every node of the graph is a bin of the final packing, the items it contains are the original items merged into it, and the number of remaining nodes is the objective value of the solution.

\begin{figure}[!h]
\centering
\shrinkbox{\scalebox{0.7}{
\begin{tikzpicture}[
    itemnode/.style={circle, fill=orange!40, draw=none, minimum size=8mm, align=center, font=\small},
    edgelabel/.style={fill=white, inner sep=1pt, font=\scriptsize},
    activeedge/.style={thick, dashed, orange}
]
\begin{scope}[xshift=0cm, yshift=6.3cm]
    \node[itemnode, label=below:{$w_1=1$}] (d1) at (1.5, 0) {$d_1$};
    \node[itemnode, label=above:{$w_2=2$}] (d2) at (1.5, 2.5) {$d_2$};
    \node[itemnode, label=above:{$w_3=4$}] (d3) at (3.5, 2.5) {$d_3$};
    \node[itemnode, label=below:{$w_4=5$}] (d4) at (3.5, 0) {$d_4$};
    \node[itemnode, label=below:{$w_5=9$}] (d5) at (0, 1.25) {$d_5$};
    \draw[thick] (d2) -- (d5) node[midway, edgelabel] {$k_1$};
    \draw[thick] (d1) -- (d5) node[midway, edgelabel] {$k_2$};
    \draw[thick] (d1) -- (d2) node[midway, edgelabel] {$k_3$};
    \draw[thick] (d2) -- (d4) node[pos=0.3, edgelabel] {$k_4$};
    \draw[thick] (d2) -- (d3) node[midway, edgelabel] {$k_5$};
    \draw[thick] (d1) -- (d3) node[pos=0.3, edgelabel] {$k_6$};
    \draw[thick] (d1) -- (d4) node[midway, edgelabel] {$k_7$};
    \draw[thick] (d3) -- (d4) node[midway, edgelabel] {$k_8$};
    \node at (2.25, -1.3) {(a) Initial state $s_0$};
\end{scope}
\begin{scope}[xshift=6cm, yshift=6.3cm]
    \node[itemnode, label=below:{$w_1=1$}] (d1) at (1.5, 0) {$d_1$};
    \node[itemnode, label=above:{$w_2=2$}] (d2) at (1.5, 2.5) {$d_2$};
    \node[itemnode, label=above:{$w_3=4$}] (d3) at (3.5, 2.5) {$d_3$};
    \node[itemnode, label=below:{$w_4=5$}] (d4) at (3.5, 0) {$d_4$};
    \node[itemnode, label=below:{$w_5=9$}] (d5) at (0, 1.25) {$d_5$};
    \draw[thick] (d2) -- (d5) node[midway, edgelabel] {$k_1$};
    \draw[thick] (d1) -- (d5) node[midway, edgelabel] {$k_2$};
    \draw[thick] (d1) -- (d2) node[midway, edgelabel] {$k_3$};
    \draw[thick] (d2) -- (d4) node[pos=0.3, edgelabel] {$k_4$};
    \draw[thick] (d2) -- (d3) node[midway, edgelabel] {$k_5$};
    \draw[thick] (d1) -- (d3) node[pos=0.3, edgelabel] {$k_6$};
    \draw[activeedge] (d1) -- (d4) node[midway, edgelabel] {$k_7$};
    \draw[thick] (d3) -- (d4) node[midway, edgelabel] {$k_8$};
    \node at (2.25, -1.3) {(b) Action $a_0 = k_7$};
\end{scope}
\begin{scope}[xshift=12cm, yshift=6.3cm]
    \node[itemnode, label=below:{$\ell_6=6$}] (d6) at (2.5, 0) {$d_6$};
    \node[itemnode, label=above:{$w_2=2$}] (d2) at (1.5, 2.5) {$d_2$};
    \node[itemnode, label=above:{$w_3=4$}] (d3) at (3.5, 2.5) {$d_3$};
    \node[itemnode, label=below:{$w_5=9$}] (d5) at (0, 1.25) {$d_5$};
    \draw[thick] (d2) -- (d5) node[midway, edgelabel] {$k_1$};
    \draw[thick] (d6) -- (d2) node[midway, edgelabel] {$k_3$};
    \draw[thick] (d2) -- (d3) node[midway, edgelabel] {$k_5$};
    \draw[thick] (d6) -- (d3) node[midway, edgelabel] {$k_6$};
    \node at (2.25, -1.3) {(c) State $s_1$};
\end{scope}
\begin{scope}[xshift=0cm, yshift=0cm]
    \node[itemnode, label=below:{$\ell_6=6$}] (d6) at (2.5, 0) {$d_6$};
    \node[itemnode, label=above:{$w_2=2$}] (d2) at (1.5, 2.5) {$d_2$};
    \node[itemnode, label=above:{$w_3=4$}] (d3) at (3.5, 2.5) {$d_3$};
    \node[itemnode, label=below:{$w_5=9$}] (d5) at (0, 1.25) {$d_5$};
    \draw[thick] (d2) -- (d5) node[midway, edgelabel] {$k_1$};
    \draw[activeedge] (d6) -- (d2) node[midway, edgelabel] {$k_3$};
    \draw[thick] (d2) -- (d3) node[midway, edgelabel] {$k_5$};
    \draw[thick] (d6) -- (d3) node[midway, edgelabel] {$k_6$};
    \node at (2.25, -1.3) {(d) Action $a_1 = k_3$};
\end{scope}
\begin{scope}[xshift=6cm, yshift=0cm]
    \node[itemnode, label=above:{$\ell_7=8$}] (d7) at (2, 2.5) {$d_7$};
    \node[itemnode, label=above:{$w_3=4$}] (d3) at (4, 2.5) {$d_3$};
    \node[itemnode, label=below:{$w_5=9$}] (d5) at (0, 1.25) {$d_5$};
    \node at (2.25, -1.3) {(e) Terminal state $s_T$};
\end{scope}
\end{tikzpicture}
}}
\caption{One episode of the MDP on the example instance. (a) Initial state. (b) Selection of $a_0$. (c) Graph after the first merge; edges of the new node are recomputed from its load. (d) Selection of $a_1$. (e) Terminal state with three isolated nodes, i.e.\ three bins.}
\label{fig:example_mdp}
\end{figure}
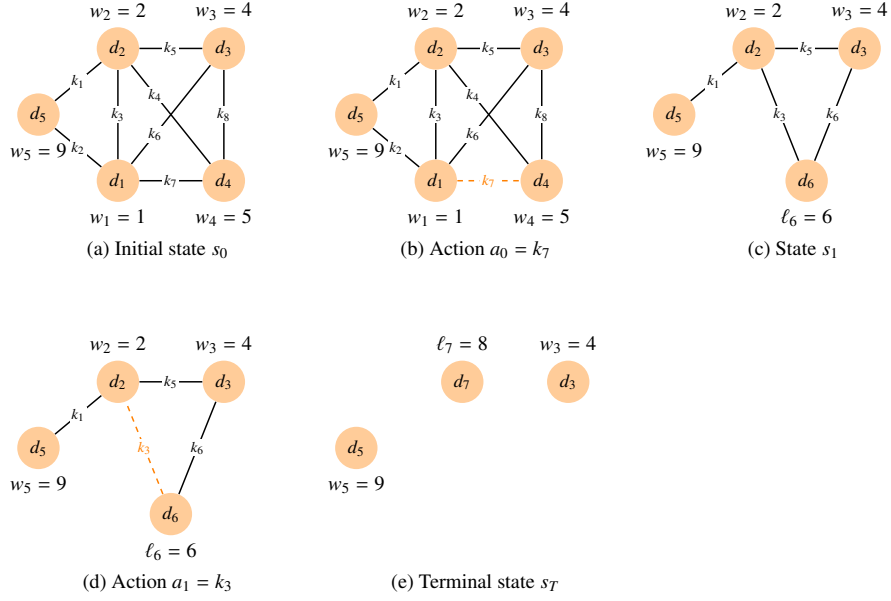

\paragraph{Illustrative example (continued)} Figure~\ref{fig:example_mdp} traces one episode of the MDP on the five-item instance of Figure~\ref{fig:example_graph}. The initial state $s_0$ (Figure~\ref{fig:example_mdp}a) is the item-compatibility graph itself, so all eight edges are available as actions. Suppose the policy selects $a_0=k_7$ (Figure~\ref{fig:example_mdp}b), the edge between $d_1$ and $d_4$. The two items, of weights 1 and 5, are merged into a new node $d_6$ of load 6, and the edges of $d_6$ are recomputed from this load. Only $d_2$ and $d_3$ still fit together with $d_6$, whereas $d_5$ does not (Figure~\ref{fig:example_mdp}c). The policy next selects $a_1=k_3$ (Figure~\ref{fig:example_mdp}d) and merges the item of weight 2 into $d_6$, producing $d_7$ of load 8. Now no two remaining nodes fit together, since every pairwise sum exceeds the capacity. The episode ends in the terminal state $s_T$ with the three isolated nodes $d_3$, $d_5$ and $d_7$ (Figure~\ref{fig:example_mdp}e), which are the three bins of the solution, $\{4\}$, $\{9\}$ and $\{1,5,2\}$. Two merges were made, so the return is $1$ and the number of bins is $5-2=3$. This solution is not optimal. Selecting $k_1$ first (items 2 and 9) and then packing items 1, 4 and 5 together reaches the optimum of two bins, which shows that the order of merges matters and is exactly what the policy has to learn.

\subsection{DRL architecture}
\label{sec:method:arch}
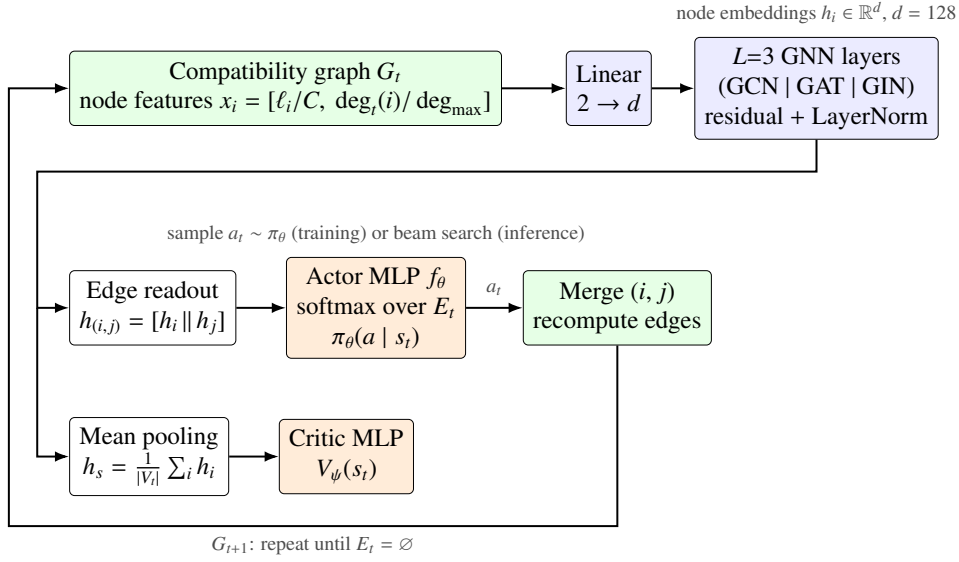
\begin{figure}[!t]
\centering
\shrinkbox{%
\begin{tikzpicture}[
  font=\small,
  box/.style={draw, rounded corners=2pt, align=center, minimum height=1.05cm, inner sep=4pt, fill=white},
  enc/.style={box, fill=blue!8},
  head/.style={box, fill=orange!15},
  env/.style={box, fill=green!10},
  arr/.style={-{Latex[length=2mm]}, thick},
  lab/.style={font=\scriptsize, text=black!70}
]
\node[env] (graph) {Compatibility graph $G_t$\\ node features $x_i=[\ell_i/C,\ \deg_t(i)/\deg_{\max}]$};
\node[enc, right=0.9cm of graph] (embed) {Linear\\ $2\to d$};
\node[enc, right=0.6cm of embed] (gnn) {$L{=}3$ GNN layers\\ (GCN $|$ GAT $|$ GIN)\\ residual + LayerNorm};
\node[lab, above=0.05cm of gnn] {node embeddings $h_i\in\mathbb{R}^{d}$, $d=128$};
\node[box, anchor=west] (edge) at ([yshift=-3.1cm]graph.west) {Edge readout\\ $h_{(i,j)}=[h_i\,\|\,h_j]$};
\node[box, anchor=west] (pool) at ([yshift=-5.2cm]graph.west) {Mean pooling\\ $h_s=\frac{1}{|V_t|}\sum_i h_i$};
\node[head, right=0.7cm of edge] (actor) {Actor MLP $f_\theta$\\ softmax over $E_t$\\ $\pi_\theta(a\mid s_t)$};
\node[head, right=0.7cm of pool] (critic) {Critic MLP\\ $V_\psi(s_t)$};
\node[env, right=0.8cm of actor] (step) {Merge $(i,j)$\\ recompute edges};
\node[lab, above=0.05cm of actor] {sample $a_t\sim\pi_\theta$ (training) or beam search (inference)};
\draw[arr] (graph) -- (embed);
\draw[arr] (embed) -- (gnn);
\coordinate (bus) at ([xshift=-0.45cm]edge.west);
\draw[arr] (gnn.south) -- ++(0,-0.45) -| (bus) -- (edge.west);
\draw[arr] (bus) |- (pool.west);
\draw[arr] (edge) -- (actor);
\draw[arr] (pool) -- (critic);
\draw[arr] (actor) -- node[above, lab, yshift=1pt]{$a_t$} (step);
\draw[arr] (step.south) -- ++(0,-2.55) -| node[pos=0.25, below, lab]{$G_{t+1}$: repeat until $E_t=\varnothing$} ([xshift=-0.85cm]graph.west) -- (graph.west);
\end{tikzpicture}%
}
\caption{Proposed actor-critic architecture. The encoder maps the current compatibility graph to node embeddings. Two size-independent readouts feed the actor and the critic. The chosen merge updates the graph and the loop repeats until no compatible pair remains.}
\label{fig:arch}
\end{figure}

Policy and value are parameterized by a GNN operating on $G_t$ (Figure~\ref{fig:arch}).

\paragraph{Graph encoder} Node features~\eqref{eq:features} are linearly embedded into $\mathbb{R}^d$ ($d=128$) and refined by $L=3$ message-passing layers, each followed by a residual connection and layer normalization, $h_i^{(l+1)}=\mathrm{LN}\big(h_i^{(l)}+\tilde h_i^{(l+1)}\big)$. We instantiate three standard aggregation schemes for $\tilde h^{(l+1)}_i$, with $\mathcal{N}(i)$ the neighbourhood of $i$ in $G_t$ and $W^{(l)}$ a learnable matrix. The graph convolutional network (GCN)~\cite{kipf2017gcn} uses isotropic, degree-normalized aggregation,
\begin{equation}
\tilde h_i^{(l+1)} = \mathrm{ReLU}\Big(\textstyle\sum_{j\in\mathcal{N}(i)\cup\{i\}} \frac{1}{\sqrt{\tilde d_i\tilde d_j}}\, W^{(l)} h_j^{(l)}\Big),
\end{equation}
where $\tilde d_i$ is the degree of $i$ including the self-loop. The graph attention network (GAT)~\cite{velickovic2018gat} replaces the fixed normalization by learnable anisotropic weights, here with $K=4$ attention heads,
\begin{equation}
\tilde h_i^{(l+1)} = \mathrm{ReLU}\Big(\textstyle\sum_{j\in\mathcal{N}(i)\cup\{i\}} \alpha_{ij}^{(l)}\, W^{(l)} h_j^{(l)}\Big),
\end{equation}
where $\alpha_{ij}^{(l)}$ is a softmax over $\mathcal{N}(i)\cup\{i\}$ of a single-layer scoring network applied to $[W^{(l)}h_i^{(l)}\,\|\,W^{(l)}h_j^{(l)}]$. The graph isomorphism network (GIN)~\cite{xu2019gin} uses sum aggregation followed by an MLP update,
\begin{equation}
\tilde h_i^{(l+1)} = \mathrm{MLP}^{(l)}\Big((1+\epsilon^{(l)})\,h_i^{(l)} + \textstyle\sum_{j\in\mathcal{N}(i)} h_j^{(l)}\Big),
\end{equation}
where $\epsilon^{(l)}$ is a learnable scalar.

\paragraph{Size-independent readouts} From the final embeddings $h_i=h_i^{(L)}$ two representations are formed. The \emph{state} vector is the mean of the node embeddings, $h_s=\frac{1}{|V_t|}\sum_{i\in V_t}h_i$, and the \emph{action} vector of a candidate edge is the concatenation of its endpoints, $h_{(i,j)}=[h_i\,\|\,h_j]\in\mathbb{R}^{2d}$. Both have fixed dimension regardless of $|V_t|$ and $|E_t|$, so a single trained network applies to graphs of any size.

\paragraph{Actor and critic} The actor scores every feasible edge with a three-layer MLP $f_\theta:\mathbb{R}^{2d}\to\mathbb{R}$ (hidden widths 128 and 64) and normalizes over the current edge set,
\begin{equation}
\pi_\theta(a=(i,j)\mid s_t)=\frac{\exp f_\theta(h_{(i,j)})}{\sum_{(u,v)\in E_t}\exp f_\theta(h_{(u,v)})}.
\label{eq:policy}
\end{equation}
The critic predicts the state value $V_\psi(s_t)$ from $h_s$ with an MLP of the same shape. For SAC and the value-based algorithms, the same action vectors feed a $Q$-network $Q_\phi(s_t,a)$ of the same shape, and the value-based policies select $\arg\max_{a\in E_t}Q_\phi(s_t,a)$.

\subsection{Training}
\label{sec:method:train}
The MDP and the architecture do not depend on the learning algorithm. We train policies with three policy-gradient methods, REINFORCE~\cite{williams1992reinforce}, Advantage Actor--Critic (A2C)~\cite{mnih2016a3c} and Proximal Policy Optimization (PPO)~\cite{schulman2017ppo}, which optimize $\pi_\theta$ directly and use the critic as a baseline. We also train with Soft Actor--Critic (SAC)~\cite{haarnoja2018sac}, a maximum-entropy actor--critic method that learns a soft $Q$-function and a stochastic policy, and with two value-based methods, Deep Q-Network (DQN)~\cite{mnih2015dqn} and SARSA~\cite{rummery1994sarsa}, which learn $Q_\phi$ only and act greedily on it with $\epsilon$-greedy exploration during training. All follow their standard formulations. As the primary configuration, PPO is described here. Each training epoch samples a batch of $M$ independent instances from the training distribution, rolls out one episode per instance with the current policy, and stores every transition of these episodes in a buffer $\mathcal{B}$. For each transition it computes the reward-to-go $\hat R_t=\sum_{t'\ge t}R_{t'}$ (undiscounted, $\gamma=1$, since the horizon is finite) and the advantage $\hat A_t=\hat R_t-V_\psi(s_t)$, normalized to zero mean and unit variance over $\mathcal{B}$. The parameters are then updated for $K$ inner epochs over $\mathcal{B}$ on the clipped surrogate
\begin{equation}
\begin{split}
\mathcal{L}={}&\mathbb{E}_t\Big[\min\big(\rho_t\hat A_t,\ \mathrm{clip}(\rho_t,1-\varepsilon,1+\varepsilon)\hat A_t\big)\Big]\\
&-c_v\,\mathbb{E}_t\big[(V_\psi(s_t)-\hat R_t)^2\big]+c_e\,\mathbb{E}_t\big[\mathcal{H}(\pi_\theta(\cdot\mid s_t))\big],
\end{split}
\label{eq:ppo}
\end{equation}
with probability ratio $\rho_t=\pi_\theta(a_t\mid s_t)/\pi_{\theta_{\text{old}}}(a_t\mid s_t)$, clipping parameter $\varepsilon$, value coefficient $c_v$ and entropy coefficient $c_e$. The checkpoint with the best performance on a held-out validation set drawn from the training distribution is retained. Algorithm~\ref{alg:train} summarizes the procedure. The values of all hyperparameters are listed in Table~\ref{tab:hparams}.

\begin{algorithm}[htbp]
\caption{Training with PPO}
\label{alg:train}
\begin{algorithmic}[1]
\Require training distribution $\mathcal{D}$, epochs $T$, batch size $M$, inner epochs $K$
\State initialize encoder, actor $\theta$, critic $\psi$
\For{epoch $=1,\dots,T$}
    \State $\mathcal{B}\leftarrow\emptyset$ \Comment{buffer of transitions collected in this epoch}
    \For{$m=1,\dots,M$}
        \State sample $(W,C)\sim\mathcal{D}$; build $G_0$ by~\eqref{eq:compat}
        \While{$E_t\neq\varnothing$}
            \State $a_t\sim\pi_\theta(\cdot\mid G_t)$ by~\eqref{eq:policy}; merge $a_t$; observe $R_t=R(s_t,a_t)$
        \EndWhile
        \State $\hat R_t\leftarrow\sum_{t'\ge t}R_{t'}$ for every step $t$ of the episode
        \State append $\{(s_t,a_t,\log\pi_\theta(a_t\mid s_t),\hat R_t)\}_t$ to $\mathcal{B}$
    \EndFor
    \State $\hat A_t\leftarrow \hat R_t-V_\psi(s_t)$, normalized over $\mathcal{B}$
    \For{$k=1,\dots,K$}
        \State update $(\theta,\psi)$ by Adam on~\eqref{eq:ppo} with gradient-norm clipping
    \EndFor
    \If{validation epoch} evaluate on the validation set; keep the best checkpoint \EndIf
\EndFor
\end{algorithmic}
\end{algorithm}

\subsection{Inference: stochastic beam search}
\label{sec:method:decode}
At inference the trained policy is decoded with \emph{stochastic beam search} of width $B$ (Algorithm~\ref{alg:beam}), which keeps $B$ partial solutions in parallel. At every step, each of them samples $\min(B,|E_t|)$ distinct edges from $\pi_\theta$ without replacement. The resulting children are ranked first by their current bin count and then by cumulative log-probability, and the best $B$ are retained. Completed solutions are kept aside and the one with the fewest bins is returned. Sampling rather than taking the $B$ most probable edges lets the search reach merges to which the policy assigns low but non-negligible probability. Since every beam call is seeded, the decoder is deterministic for a given seed.

\begin{algorithm}[htbp]
\caption{Inference with stochastic beam search}
\label{alg:beam}
\begin{algorithmic}[1]
\Require items $W$, capacity $C$, trained policy $\pi_\theta$, beam width $B$
\State build $G_0$ by~\eqref{eq:compat}; $\mathcal{P}\leftarrow\{(G_0,\,0)\}$; $\mathcal{F}\leftarrow\emptyset$
\While{$\mathcal{P}\neq\emptyset$}
    \State $\mathcal{C}\leftarrow\emptyset$
    \For{each $(G,\sigma)\in\mathcal{P}$}
        \State compute $x_i$ by~\eqref{eq:features}; $\pi\leftarrow\pi_\theta(\cdot\mid G)$ by~\eqref{eq:policy}
        \State sample $k=\min(B,|E|)$ distinct edges $a^{(1)},\dots,a^{(k)}\sim\pi$ without replacement
        \For{$r=1,\dots,k$}
            \State $G'\leftarrow$ merge $a^{(r)}$ in $G$; $\sigma'\leftarrow\sigma+\log\pi(a^{(r)})$
            \If{$E(G')=\varnothing$} add $G'$ to $\mathcal{F}$ \Else{} add $(G',\sigma')$ to $\mathcal{C}$ \EndIf
        \EndFor
    \EndFor
    \State $\mathcal{P}\leftarrow$ the $B$ elements of $\mathcal{C}$ with fewest nodes $|V|$, ties broken by largest $\sigma'$
\EndWhile
\State \Return $\arg\min_{G\in\mathcal{F}}|V(G)|$ \Comment{each remaining node is a bin}
\end{algorithmic}
\end{algorithm}

\section{Experiments and Results}
\label{sec:experiments}

\subsection{Benchmark instances}
\label{sec:exp:data}
We use the BPPLIB library~\cite{delorme2018bpplib} and evaluate on \emph{all} instances of its nine complete families (Table~\ref{tab:bpplib_stats}). Optimal or best-known values are taken from BPPLIB. The families differ widely in item count, capacity and weight structure and are therefore out of distribution for a model trained on uniform 50-item instances. 

\begin{table*}[htbp]
\centering
\caption{BPPLIB families}
\label{tab:bpplib_stats}
\footnotesize
\begin{tabular}{lrll>{\raggedright\arraybackslash}p{3.1cm}}
\toprule
\textbf{Family} & \textbf{Instances} & \textbf{Items $n$} & \textbf{Capacity $C$} & \textbf{Weight distribution} \\
\midrule
Falkenauer T & 80 & 60, 120, 249, 501 & 1,000 & triplets, $w\in[250,500]$, each bin filled exactly by three items \\
Falkenauer U & 80 & 120, 250, 500, 1,000 & 150 & uniform $[20,100]$ \\
Scholl 1 & 720 & 50, 100, 200, 500 & 100, 120, 150 & uniform $[1,100]$, $[20,100]$, $[30,100]$ \\
Scholl 2 & 480 & 50, 100, 200, 500 & 1,000 & uniform, average of 3, 5, 7 or 9 items per bin \\
Scholl 3 & 10 & 200 & 100,000 & uniform $[20{,}000,35{,}000]$ \\
Schwerin 1 & 100 & 100 & 1,000 & uniform $[150,200]$ \\
Schwerin 2 & 100 & 120 & 1,000 & uniform $[150,200]$ \\
W\"ascher & 17 & 57--239 & 10,000 & heterogeneous item classes \\
Hard28 & 28 & 160, 180, 200 & 1,000 & instances selected to be hard for exact methods \\
\midrule
Total & 1,615 & & & \\
\bottomrule
\end{tabular}
\end{table*}

\subsection{Baselines and comparators}
\label{sec:exp:baselines}
\paragraph{Constructive heuristic} FFD~\cite{johnson1974} is the reference constructive heuristic. BFD produced identical bin counts on our validation data and is omitted from the tables.

\paragraph{Grouping metaheuristic} We implemented a GGA following Falkenauer~\cite{falkenauer1996hybrid}. The population contains 50 packings and is initialized with the FFD packing, the BFD packing and 48 first-fit packings of random item permutations. Parents are chosen by tournament selection of size 3. The grouping crossover keeps the fullest half of the bins of the first parent and reinserts the remaining items by BFD, following the bin order of the second parent. Two mutation operators, an item swap between two bins and an item move to another bin, are applied with probability 0.3, and the best 10\% of the population is carried over unchanged. Each run is limited to 2\,s per instance, or at most 500 generations. 

\paragraph{Learned solvers} We compare our work with the published results of Shi et al.~\cite{shi2025combination}, who report all nine BPPLIB families. The comparison includes their bipartite-GCN method with column generation, with and without the monotonicity cut (BGCN, BGCNMC), and the three RL baselines they report. These baselines are pointer-network RL (PTR), hierarchical RL with graph pointer networks (HRL-GPN) and ranked-reward MCTS (RRMCTS).

\subsection{Metrics and evaluation protocol}
\label{sec:exp:protocol}
Following~\cite{shi2025combination} we report the \emph{per-instance optimality gap} $(b/\mathrm{opt}-1)\times 100\%$, where $b$ is the number of bins a method uses on an instance and $\mathrm{opt}$ is the optimal (or best-known) number of bins of that instance from BPPLIB~\cite{delorme2018bpplib}. For each family we report the mean and the standard deviation of this gap over its instances. We also report the number of instances solved to optimality and the numbers of instances on which the proposed solver uses fewer or more bins than FFD. Gaps are computed against the fixed BPPLIB optima and are hence independent of hardware. Since beam decoding is stochastic, every instance is solved with three decoding seeds $\{0,1,2\}$. The reported gap of the proposed solver is that of the seed-averaged bin count, and optimum counts refer to seed 0 unless stated otherwise.

\subsection{Implementation details}
\label{sec:exp:impl}
All models are trained on synthetic instances with $n=50$ items, $C=100$ and weights drawn uniformly from $\{1,\dots,100\}$, for 2,000 epochs of 16 episodes with a single training seed (42). Table~\ref{tab:hparams} lists the hyperparameters.

\begin{table}[htbp]
\centering
\caption{Hyperparameters of the primary configuration.}
\label{tab:hparams}
\small
\begin{tabular}{@{}l >{\raggedright\arraybackslash}p{9.5cm}@{}}
\toprule
\textbf{Component} & \textbf{Setting} \\
\midrule
Encoder & GCN, $L=3$ layers, $d=128$, residual + LayerNorm, dropout 0.1 \\
Actor / critic MLP & $2d\to128\to64\to1$ / $d\to128\to64\to1$ \\
Algorithm & PPO, $\varepsilon=0.2$, $c_v=0.5$, $c_e=0.01$, $K=4$ inner epochs \\
Optimizer & Adam, learning rate $3\times10^{-4}$, gradient-norm clip 1.0 \\
Discount & $\gamma=1$ (finite horizon) \\
Training data & $n=50$, $C=100$, $w_i\sim U\{1,\dots,100\}$ \\
Budget & 2,000 epochs $\times$ 16 episodes; seed 42 \\
Validation & 20 fixed instances of the training distribution, every 50 epochs; best checkpoint retained \\
Decoding & stochastic beam search, $B=5$, seeds $\{0,1,2\}$ \\
Hardware & Intel Core i5-13500H (2.60\,GHz), 32\,GB RAM \\
\bottomrule
\end{tabular}
\end{table}

\subsection{Design study: encoder and learning algorithm}
\label{sec:exp:design}
We trained all 18 encoder$\times$algorithm combinations (GCN, GAT, GIN $\times$ PPO, A2C, REINFORCE, DQN, SAC, SARSA) with identical budgets and compared the selected checkpoint of each on the 20 in-distribution validation instances, decoded with the stochastic beam search used in all experiments (Table~\ref{tab:validation}). DQN and SARSA learn action values only and have no sampling distribution, so beam search does not apply to them. Under greedy decoding they remain far behind the other models (27.10-30.05 bins) and are not considered further. On these uniform 50-item instances FFD, BFD and the 2\,s GGA all use 26.20 bins on average against a lower bound $L_1$ of 25.15, so the validation set leaves almost no headroom and 26.20 is the level of the strongest baselines. Two observations follow. First, only the GCN and GAT PPO models reach this level, with zero variance across seeds, whereas the REINFORCE, A2C and SAC models end 0.07-1.03 bins above it. Their policies are less peaked, so the alternatives that the beam samples are more often worse than the most probable merge, and the beam returns them when they tie on bin count. PPO is therefore the best learning algorithm. Second, GCN+PPO and GAT+PPO tie. We select GCN for its simpler, parameter-free aggregation and its smoother training curve. This GCN+PPO model is the proposed solver in all remaining experiments and the tables label it ``Ours''.

The choice of encoder is also confirmed on the benchmark. A beam-decoded grid on small subsets of Falkenauer~T and Scholl~2 placed the GIN models ahead of all others, with GIN+PPO first on Scholl~2 and third on Falkenauer~T (Appendix~A, Table~\ref{tab:fullbeam}). On the benchmark, however, GIN+PPO trailed GCN+PPO on most of the families. The choice made on in-distribution validation data therefore agrees with full-benchmark generalization. Selection on a narrow subset would have been misleading.

\begin{table}[htbp]
\centering
\caption{Design study: average bins of the encoder$\times$algorithm combination on the 20 in-distribution validation instances ($B=5$, mean $\pm$ std over three seeds).  References on the same instances: FFD = BFD = GGA (2\,s) 26.20, lower bound $L_1$ 25.15. Best value in bold.}
\label{tab:validation}
\small
\shrinkbox{%
\begin{tabular}{lcccc}
\toprule
\textbf{Encoder} & PPO & REINFORCE & A2C & SAC \\
\midrule
GCN & \textbf{26.20} $\pm$ 0.00 & 26.82 $\pm$ 0.06 & 26.27 $\pm$ 0.02 & 27.20 $\pm$ 0.04 \\
GAT & \textbf{26.20} $\pm$ 0.00 & 26.52 $\pm$ 0.02 & 26.85 $\pm$ 0.11 & 26.52 $\pm$ 0.02 \\
GIN & 26.37 $\pm$ 0.06 & 26.35 $\pm$ 0.04 & 27.23 $\pm$ 0.06 & 26.27 $\pm$ 0.02 \\
\bottomrule
\end{tabular}%
}
\end{table}

\subsection{Results}
\label{sec:exp:main}
Table~\ref{tab:main_results} reports the main result of the uniform-trained GCN+PPO model, decoded with $B=5$ and three seeds, on all 1,615 instances. The proposed policy lowers the mean gap of FFD from 2.66\% to 2.31\% and uses fewer bins than FFD on 291 instances against 170 on which it uses more (1,154 ties). The improvement is concentrated on the high-headroom families. On Falkenauer~T the gap drops from 14.69\% to 11.88\% with fewer bins on 75 of 80 instances and none lost. On Scholl~3 it drops from 6.06\% to 4.40\% on all 10 instances, and on Scholl~2 from 3.01\% to 2.31\% (119 wins, 54 losses). On Falkenauer~U and Schwerin~2 the gains are small but positive, but on Schwerin~1 and W\"ascher the proposed solver coincides with FFD on every instance. On the low-headroom Hard28 and Scholl~1 families it is marginally worse than FFD (1.21\% vs 1.20\%, and 0.51\% vs 0.47\%), losing 101 Scholl~1 instances by one bin while winning 54. In summary, the proposed method improves on FFD in five families, ties in two and is slightly behind in two. The two sides of this comparison are not symmetric. Where the proposed policy loses to FFD, it does so by a single bin (by more than one bin on only 2 of the 1,615 instances), whereas where it wins it saves up to 12 bins, and at least two bins on 76 instances. Over the whole benchmark it uses 128,702 bins against 129,080 for FFD.

The per-instance results are stable across decoding seeds. The spread of the family mean gap over the three seeds is at most 0.46 percentage points (Schwerin~2) and below 0.2 points on seven of nine families. By count of optimal solutions, however, FFD (795 of 1,615) is ahead of the proposed solver (729; 721 and 748 for the other two seeds). The benchmark is dominated by easy Scholl~1 instances on which FFD is already optimal and the proposed solver occasionally spends one extra bin. The GGA remains clearly ahead of both on every family (0.51\% mean gap, 1,243 optima).

\begin{table*}[htbp]
\centering
\caption{Zero-shot results of the proposed solver (Ours) on all nine BPPLIB families. Gap: per-instance optimality gap (\%), mean (for Ours, $\pm$ standard deviation over instances); lower is better. ``$<$FFD'' / ``$>$FFD'': instances on which Ours uses fewer / more bins than FFD. Optima: instances solved to the BPPLIB optimum. Best of FFD and Ours per family in bold; GGA shown for reference.}
\label{tab:main_results}
\small
\shrinkbox{%
\begin{tabular}{lrrrrrrrrr}
\toprule
 & & \multicolumn{3}{c}{Gap (\%)} & \multicolumn{2}{c}{Ours vs FFD} & \multicolumn{3}{c}{Optima} \\
\cmidrule(lr){3-5}\cmidrule(lr){6-7}\cmidrule(lr){8-10}
\textbf{Family} & $n$ & FFD & GGA & Ours & $<$FFD & $>$FFD & FFD & GGA & Ours \\
\midrule
Falkenauer T & 80 & 14.69 & 4.06 & \textbf{11.88} $\pm$ 1.42 & 75 & 0 & 0 & 0 & 0 \\
Scholl 3 & 10 & 6.06 & 0.36 & \textbf{4.40} $\pm$ 0.81 & 10 & 0 & 0 & 8 & 0 \\
Schwerin 1 & 100 & 5.56 & 0.22 & 5.56 $\pm$ 0.00 & 0 & 0 & 0 & 96 & 0 \\
W\"ascher & 17 & 5.45 & 2.78 & 5.45 $\pm$ 2.57 & 0 & 0 & 2 & 8 & 2 \\
Schwerin 2 & 100 & 4.90 & 0.10 & \textbf{4.82} $\pm$ 1.23 & 9 & 2 & 0 & 98 & 0 \\
Scholl 2 & 480 & 3.01 & 0.47 & \textbf{2.31} $\pm$ 2.78 & 119 & 54 & 236 & 375 & 222 \\
Falkenauer U & 80 & 1.37 & 0.74 & \textbf{1.28} $\pm$ 0.54 & 24 & 12 & 6 & 26 & 9 \\
Hard28 & 28 & \textbf{1.20} & 1.20 & 1.21 $\pm$ 0.58 & 0 & 1 & 5 & 5 & 5 \\
Scholl 1 & 720 & \textbf{0.47} & 0.14 & 0.51 $\pm$ 0.96 & 54 & 101 & 546 & 627 & 491 \\
\midrule
\textbf{Total} & \textbf{1,615} & 2.66 & 0.51 & \textbf{2.31} & 291 & 170 & 795 & 1,243 & 729 \\
\bottomrule
\end{tabular}%
}
\end{table*}

Since the model is trained only on 50-item instances, Table~\ref{tab:size} breaks the four largest families down by item count. Two observations follow. First, generalization to instances up to twenty times larger than the training size does not degrade. On Falkenauer~T the improvement over FFD is robustly maintained across all sizes, at 2.6--2.9 percentage points (13.08\% vs 16.00\% at $n=60$; 10.98\% vs 13.80\% at $n=501$), with fewer bins on all 20 instances for $n\ge120$, and on Scholl~2 the gap of the proposed solver stays between 2.2\% and 2.5\% at every size while the number of instances won over FFD rises from 20 to 48. Second, on the low-headroom Scholl~1 family the balance tips towards FFD as $n$ grows (24 wins against 67 losses at $n=500$). When almost every instance is FFD-optimal, longer episodes give the stochastic decoder more opportunities to spend one extra bin. On Falkenauer~U the proposed solver is slightly ahead at every size.

\begin{table}[t]
\centering
\caption{Results by item count $n$ for the four largest families (per-instance gap \%. Ours: proposed solver). ``$<$'' / ``$>$'': instances on which Ours uses fewer / more bins than FFD.}
\label{tab:size}
\footnotesize
\shrinkbox{%
\begin{tabular}{llrrrrrr}
\toprule
\textbf{Family} & $n$ & Inst. & mean opt & FFD & GGA & Ours & $<$ / $>$ FFD \\
\midrule
\multirow{4}{*}{Falkenauer T} & 60 & 20 & 20.0 & 16.00 & 5.00 & \textbf{13.08} & 15 / 0 \\
 & 120 & 20 & 40.0 & 14.50 & 2.50 & \textbf{11.92} & 20 / 0 \\
 & 249 & 20 & 83.0 & 14.46 & 3.49 & \textbf{11.53} & 20 / 0 \\
 & 501 & 20 & 167.0 & 13.80 & 5.24 & \textbf{10.98} & 20 / 0 \\
\midrule
\multirow{4}{*}{Scholl 2} & 50 & 120 & 10.3 & 2.97 & 0.05 & \textbf{2.23} & 20 / 3 \\
 & 100 & 120 & 20.1 & 2.80 & 0.29 & \textbf{2.24} & 20 / 8 \\
 & 200 & 120 & 39.6 & 3.26 & 0.48 & \textbf{2.52} & 31 / 13 \\
 & 500 & 120 & 98.7 & 3.03 & 1.08 & \textbf{2.26} & 48 / 30 \\
\midrule
\multirow{4}{*}{Falkenauer U} & 120 & 20 & 49.1 & 1.44 & 0.41 & \textbf{1.34} & 3 / 1 \\
 & 250 & 20 & 101.6 & 1.48 & 0.49 & \textbf{1.41} & 3 / 2 \\
 & 500 & 20 & 201.2 & 1.34 & 0.87 & \textbf{1.24} & 8 / 5 \\
 & 1,000 & 20 & 400.6 & 1.21 & 1.20 & \textbf{1.14} & 10 / 4 \\
\midrule
\multirow{4}{*}{Scholl 1} & 50 & 180 & 26.6 & 0.56 & 0.02 & \textbf{0.53} & 9 / 5 \\
 & 100 & 180 & 52.1 & \textbf{0.44} & 0.09 & 0.49 & 9 / 11 \\
 & 200 & 180 & 102.7 & \textbf{0.49} & 0.16 & 0.54 & 12 / 18 \\
 & 500 & 180 & 254.1 & \textbf{0.40} & 0.28 & 0.47 & 24 / 67 \\
\bottomrule
\end{tabular}%
}
\end{table}

Table~\ref{tab:learned} compares the proposed solver with the learned methods of Shi et al.~\cite{shi2025combination}, who evaluate all nine BPPLIB families used here with the identical per-instance gap and deterministic decoding of their policies. Three findings follow. First, the pointer-network baseline PTR essentially reproduces FFD. It equals FFD on six families and is behind it by at most 0.4 points on the other three, so our solver attains a lower gap than PTR on six families, ties on Schwerin~1 and W\"ascher, and is higher only on Hard28 by 0.01 points. Second, our solver is far more stable across distributions than HRL-GPN and RRMCTS. Those methods reach near-zero gaps on the Schwerin families but degrade to 13--14\% on Scholl~3, 18--19\% on Falkenauer~T, 27--28\% on Scholl~1 and 38--40\% on Falkenauer~U and Hard28, whereas the largest gap of our solver is 11.88\% and it stays below 1.3\% on Falkenauer~U, Hard28 and Scholl~1. It attains a lower gap than both on seven of nine families. Third, against BGCN/BGCNMC, which solve linear and integer programs inside a column-generation loop at inference time, the proposed method is ahead on the three families with the most structure to exploit (Falkenauer~T 11.88\% vs 12.57\%, Scholl~2 2.31\% vs 2.36--2.38\%, Falkenauer~U 1.28\% vs 1.34\%), ties on W\"ascher, and is behind on Scholl~3 and the Schwerin families, where column generation is particularly effective, and by small margins on Hard28 and Scholl~1.

\begin{table*}[t]
\centering
\caption{Comparison with the learned solvers of Shi et al.~\cite{shi2025combination} on the nine BPPLIB families (per-instance optimality gap \%, lower is better; their values from Table~II of~\cite{shi2025combination}). BGCN/BGCNMC: bipartite GCN with column generation, without / with monotonicity cut. PTR, HRL-GPN, RRMCTS: their RL baselines. Best per family in bold, the multi-way ties on W\"ascher and Hard28 are left unbolded.}
\label{tab:learned}
\small
\shrinkbox{%
\begin{tabular}{lrrrrrrr}
\toprule
 & \multicolumn{2}{c}{This work / heuristic} & \multicolumn{2}{c}{Shi et al.\ (proposed)} & \multicolumn{3}{c}{Shi et al.\ (RL baselines)} \\
\cmidrule(lr){2-3}\cmidrule(lr){4-5}\cmidrule(lr){6-8}
\textbf{Family} & Ours & FFD & BGCN & BGCNMC & PTR & HRL-GPN & RRMCTS \\
\midrule
Falkenauer T & \textbf{11.88} & 14.69 & 12.57 & 12.57 & 14.69 & 19.08 & 17.92 \\
Scholl 3 & 4.40 & 6.06 & \textbf{1.43} & \textbf{1.43} & 6.06 & 13.88 & 12.82 \\
Schwerin 1 & 5.56 & 5.56 & 4.67 & 4.67 & 5.56 & 0.39 & \textbf{0.00} \\
W\"ascher & 5.45 & 5.45 & 5.45 & 5.45 & 5.45 & 7.86 & 6.86 \\
Schwerin 2 & 4.82 & 4.90 & 2.70 & 2.70 & 4.90 & 0.38 & \textbf{0.33} \\
Scholl 2 & \textbf{2.31} & 3.01 & 2.36 & 2.38 & 3.09 & 7.04 & 6.81 \\
Falkenauer U & \textbf{1.28} & 1.37 & 1.34 & 1.34 & 1.68 & 39.53 & 39.12 \\
Hard28 & 1.21 & 1.20 & 1.20 & 1.20 & 1.20 & 38.08 & 38.19 \\
Scholl 1 & 0.51 & 0.47 & \textbf{0.43} & \textbf{0.43} & 0.85 & 27.73 & 27.43 \\
\midrule
\multicolumn{8}{l}{\footnotesize Families on which Ours attains a lower gap than: PTR 6/9 (2 ties),} \\
\multicolumn{8}{l}{\footnotesize HRL-GPN 7/9, RRMCTS 7/9, BGCN 3/9 (1 tie), BGCNMC 3/9 (1 tie).} \\
\bottomrule
\end{tabular}%
}
\end{table*}

\subsection{Ablation studies}
\label{sec:exp:ablation}

\subsubsection{Reward function}
\label{sec:exp:ablation:reward}
We compare the per-step reward of Section~\ref{sec:method:mdp} with two alternatives. The \emph{terminal} reward is $0$ at every step and equal to the negative number of bins at the end of the episode. It is objective-equivalent to the per-step reward but sparse. The \emph{shaped} reward adds a dense potential-based term to the per-step reward, $R=1+\lambda\cdot 2\ell_i\ell_j/C^2$ for merging nodes of loads $\ell_i$ and $\ell_j$. It is the increase of the potential $\Phi(s)=\sum_i(\ell_i/C)^2$ caused by the merge and therefore favours tight local packings ($\lambda=1$). Table~\ref{tab:reward} compares the per-step reward with the shaped reward on the full benchmark (GCN+PPO, three seeds, $B=5$). The shaped reward improves the single highest-headroom family, Falkenauer~T (10.47\% vs 11.88\%). However, it degrades every other family, especially the low-headroom ones (Hard28 5.49\% vs 1.21\%; Scholl~1 2.58\% vs 0.51\%), and finds only 342 optima against 729. Its overall gap (3.81\%) is worse than FFD (2.66\%). The dense shaping signal biases the policy towards tight local packings that pay off only where the instance structure rewards them. The sparse terminal reward, in turn, is never better than the per-step reward. On the 60-item Falkenauer~T instances the per-step reward wins 3 instances and loses none (22.65 vs 22.80 bins). We therefore retain the simplest, objective-equivalent per-step reward.

\begin{table}[htbp]
\centering
\caption{Reward ablation on the full benchmark (GCN+PPO, three seeds, $B=5$): per-instance gap (\%) of the per-step and shaped rewards, with FFD for reference. Better of the two rewards per family in bold.}
\label{tab:reward}
\small
\begin{tabular}{lrrrr}
\toprule
\textbf{Family} & $n$ & FFD & Per-step & Shaped \\
\midrule
Falkenauer T & 80 & 14.69 & 11.88 & \textbf{10.47} \\
Scholl 3 & 10 & 6.06 & \textbf{4.40} & 7.18 \\
Schwerin 1 & 100 & 5.56 & \textbf{5.56} & 7.48 \\
W\"ascher & 17 & 5.45 & \textbf{5.45} & 5.52 \\
Schwerin 2 & 100 & 4.90 & \textbf{4.82} & 6.09 \\
Scholl 2 & 480 & 3.01 & \textbf{2.31} & 2.99 \\
Falkenauer U & 80 & 1.37 & \textbf{1.28} & 4.20 \\
Hard28 & 28 & 1.20 & \textbf{1.21} & 5.49 \\
Scholl 1 & 720 & 0.47 & \textbf{0.51} & 2.58 \\
\midrule
Total & 1,615 & 2.66 & \textbf{2.31} & 3.81 \\
Optima (seed 0) & 1,615 & 795 & \textbf{729} & 342 \\
\bottomrule
\end{tabular}
\end{table}

\subsubsection{Training distribution}
\label{sec:exp:ablation:dist}
Since the benchmark families are out of distribution for the uniform-trained model, we tested two alternatives with the GCN+PPO configuration: (i) a \emph{mixed} curriculum in which each episode draws one of five BPPLIB-matched distributions (three uniform ranges, a capped range $U[1,0.63C]$ and a triplet-like range $U[C/4,C/2]$), and (ii) \emph{specialists} trained for 3,000 epochs on the distribution of a target family. Both were decoded with $B=5$ and a single seed and are compared with the uniform model on the same instances with the per-instance gap. Neither improved on the uniform model. The mixed model was worse on the slice evaluated (Scholl~1, $n=50$: 3.99\% vs 0.53\%). The specialists were evaluated on their complete target families and lost to the uniform model on each. The capped specialist reached 14.17\% on Scholl~2 against 2.35\%, using more bins than the uniform model on 425 of the 480 instances and fewer on none. The $U[30,100]$ specialist reached 2.07\% on the Scholl~1 W4 class against 0.55\%. The $U[1,100]$ specialist coincides with the uniform model and tied on W\"ascher. Matching the training distribution to the target family therefore does not help. The uniform distribution exposes the policy to the widest variety of local configurations.

\subsubsection{Hyperparameter tuning}
\label{sec:exp:ablation:hp}
To check that the residual gap to the metaheuristic is not a tuning artefact, we optimized the learning rate, entropy coefficient, number of GNN layers and embedding width of the GCN+PPO model with Optuna (TPE sampler, median pruner, 20 sampled trials, 500 epochs each). The objective was the $B=5$ beam bin count on a held-out validation split disjoint from both the training and the test instances. The default configuration was included as a reference trial under the same protocol. The best configuration (learning rate $3.3\times10^{-4}$, 4 layers, width 128, entropy 0.046) improved the validation objective by only 0.04 bins over the defaults. Retrained for 2,000 epochs and evaluated on the disjoint test split, it used 0.17 fewer bins than the default model on Falkenauer~T and 0.09 more on Scholl~2 (Table~\ref{tab:hp}). Both differences are within the run-to-run variation of identical settings, which reached 0.4 bins across four repetitions of one configuration. The number of layers had the smallest importance (about 0.05), while width, learning rate and entropy coefficient shared the remainder roughly equally. The defaults are thus near-optimal, and the gap to the metaheuristic is structural rather than a matter of tuning.

\begin{table}[htbp]
\centering
\caption{Hyperparameter tuning (GCN+PPO, Optuna, 20 sampled trials). Test split disjoint from the tuning objective: first 10 Falkenauer~T ($n=60$, optimum 20.0) and first 15 Scholl~2 ($n=50$, optimum 17.33) instances; average bins, $B=5$, three seeds.}
\label{tab:hp}
\small
\begin{tabular}{lcc}
\toprule
\textbf{Configuration} & Falkenauer T & Scholl 2 \\
\midrule
Default (3 layers) & 22.43 $\pm$ 0.17 & \textbf{18.22} $\pm$ 0.03 \\
Tuned (4 layers) & \textbf{22.27} $\pm$ 0.05 & 18.31 $\pm$ 0.08 \\
FFD & 23.10 & 18.87 \\
GGA & 21.00 & 17.33 \\
\bottomrule
\end{tabular}
\end{table}

\section{Discussion}
\label{sec:discussion}

\subsection{The role of headroom}
To understand the pattern in Table~\ref{tab:main_results}, we split the 1,615 instances into two groups according to whether FFD already finds an optimal solution. On the 795 FFD-optimal instances the proposed solver cannot win by definition. It ties on 662 and loses one bin on 133, for a mean gap of 0.21\%. On the 820 instances where FFD is suboptimal it wins 291, loses 37 and lowers the mean gap from 5.25\% to 4.35\%. The proposed solver is thus best read as a method that improves FFD where structure leaves room, in particular on the triplet-structured Falkenauer~T instances, where the benefit is maintained up to $n=501$, rather than as a uniform replacement for FFD. The two families on which it reproduces FFD exactly, Schwerin~1 and W\"ascher, are instructive. On Schwerin~1 the items lie in $[150,200]$ with $C=1{,}000$, and on W\"ascher 99.6\% of the item pairs are compatible, so the initial compatibility graph is complete or nearly so. The normalized degree feature is then constant for most of the episode and becomes informative only when partial bins approach the capacity. The state carries little structural information beyond the loads, and the policy has little to exploit.

\subsection{Relation to metaheuristics and hybrid learned solvers}
A grouping genetic algorithm remains ahead of the proposed method on every family (0.51\% vs 2.31\%, 1,243 vs 729 optima), and none of the levers we examined, reward, training distribution, encoder or hyperparameters, closes the last few tenths of a bin. We attribute this to the level of abstraction at which learning is applied. The GGA searches at the level of whole bins with operators that preserve good groups while our policy constructs solutions by roughly $n$ local pairwise decisions among $\Theta(n^2)$ candidates. The most competitive learned solver for 1D-BPP~\cite{shi2025combination} likewise places learning at a higher level, selecting among columns proposed by column generation. The comparison in Table~\ref{tab:learned} nevertheless shows that a pure end-to-end constructor can match or exceed a solver-augmented method on the most structured families and is considerably more robust across distributions than the hierarchical-RL and MCTS baselines, while the pointer-network baseline never moves far from FFD.

\section{Conclusions and future work}
\label{sec:conclusion}
We formulated 1D-BPP as an MDP on an item-compatibility graph and developed an end-to-end graph reinforcement learning solver whose size-invariant GNN actor--critic is trained with PPO on 50-item uniform instances and decoded with stochastic beam search. A controlled design study and ablations over rewards, training distributions and hyperparameters fixed a deliberately simple configuration, all selected on in-distribution validation data. Evaluated zero-shot on all 1,615 instances of the nine BPPLIB families with three decoding seeds, the single model lowers the mean gap of FFD from 2.66\% to 2.31\%, with gains concentrated on structured, high-headroom families, attains a lower gap than a column-generation-based learned solver on three of the nine families evaluated by both studies, and is markedly more stable across distributions than hierarchical-RL and MCTS baselines. A grouping genetic algorithm remains ahead overall, and our analysis attributes the residual gap to the pairwise, constructive level at which learning is applied rather than to tuning.

When a few extra bins are tolerable, FFD remains the method of choice, and when the best attainable quality is required, a grouping metaheuristic should be used. The proposed method sits between the two. It improves FFD on structured instances with a single trained policy, without a per-instance search and without hand-designed crossover, mutation or repair operators. Its only problem-specific component is the compatibility rule that defines the edges, so the framework can be adapted to constrained variants of the problem by changing that rule alone.

This flexibility comes at a computational cost. Although the proposed size-agnostic GNN policy achieves competitive results without requiring hand-crafted rules, its inference time, particularly when decoded with stochastic beam search, is currently higher than that of highly optimized heuristic implementations such as FFD.

This work shows that a new formulation, an MDP on the item-compatibility graph solved by a single size-invariant policy, is viable for 1D-BPP. The configuration studied here was kept deliberately simple, and both the formulation and the architecture leave room for improvement. Future work may proceed in three directions. The first is to raise the level at which learning is applied, by letting the policy either select the next item for an open bin, which reduces the action space from quadratic to linear in the number of items, or construct whole bins as single actions. The second is to close the computational gap to optimized heuristic implementations by accelerating the graph update mechanism or developing more efficient greedy decoding strategies. Finally, from a knowledge-engineering perspective, future research could extend this graph-based DRL framework to dynamic or stochastic environments, such as real-time cloud resource allocation or dynamic freight loading. In such settings, where item-compatibility graphs are continuously updated with streaming data, the proposed framework has the potential to evolve into a highly adaptive, real-time intelligent decision support system.

\section*{Data availability}
The BPPLIB instances are publicly available~\cite{delorme2018bpplib}. All other materials are available upon request.

\section*{Acknowledgements}
This work is supported by the Scientific and Technological Research Council of T\"urkiye (T\"UB\.ITAK) under Grant No.\ 124M974.

\appendix
\section{Full design-study grids}
\label{app:grid}
Tables~\ref{tab:fullgreedy} and~\ref{tab:fullbeam} give the complete per-model grids underlying Section~\ref{sec:exp:design}. The greedy grid (six families, all trained models) shows the split between the policy-gradient methods and SAC, DQN and SARSA, and the collapse of greedy decoding to FFD. The beam grid ($B=20$, three seeds) gives per-model results on the first 20 Falkenauer~T and first 15 Scholl~2 instances, the discriminating subsets used during development. In Table~\ref{tab:fullgreedy} the first block of 18 rows are the uniform-trained (V1) models, the second block the mixed-trained (V2) models, and the last row a generalist model trained on the Scholl~1 distributions (V3).

\begin{table}[htbp]\centering
\caption{Greedy (deterministic, argmax) decoding: average bins for all trained models on the design-study evaluation set. Policy-based methods (A2C/PPO/REINFORCE) reproduce FFD almost exactly; SAC and the value-based methods (DQN/SARSA) fall below it. V1 = uniform training, V2 = mixed, V3 = generalist.}
\label{tab:fullgreedy}\scriptsize
\resizebox{!}{\ifdim\height>0.78\textheight 0.78\textheight\else\height\fi}{%
\begin{tabular}{l cccccc}
\toprule
Model & Falk\_T & Falk\_U & Schl\_1 & Schl\_2 & Hard28 & W\"asch \\
\midrule
gat\_a2c\_step & 23.20 & 49.10 & 27.47 & 18.87 & 72.80 & 19.60 \\
gat\_dqn\_step & 23.95 & 55.40 & 31.20 & 20.20 & 85.87 & 23.00 \\
gat\_ppo\_step & 23.20 & 49.10 & 27.47 & 18.87 & 72.80 & 19.60 \\
gat\_reinforce\_step & 23.20 & 49.10 & 27.47 & 18.87 & 72.80 & 19.60 \\
gat\_sac\_step & 23.20 & 49.70 & 27.93 & 18.80 & 76.20 & 19.60 \\
gat\_sarsa\_step & 24.10 & 58.20 & 31.13 & 20.33 & 90.00 & 22.40 \\
gcn\_a2c\_step & 23.20 & 49.10 & 27.47 & 18.87 & 72.93 & 19.60 \\
gcn\_dqn\_step & 24.00 & 53.60 & 30.40 & 20.33 & 87.07 & 22.10 \\
gcn\_ppo\_step & 23.20 & 49.10 & 27.47 & 18.87 & 72.80 & 19.60 \\
gcn\_reinforce\_step & 23.20 & 49.10 & 27.47 & 18.87 & 72.80 & 19.60 \\
gcn\_sac\_step & 24.25 & 52.00 & 27.73 & 20.47 & 74.33 & 22.10 \\
gcn\_sarsa\_step & 23.20 & 49.10 & 29.73 & 18.87 & 84.60 & 19.60 \\
gin\_a2c\_step & 23.20 & 49.20 & 27.47 & 18.93 & 73.20 & 19.60 \\
gin\_dqn\_step & 23.20 & 56.70 & 28.60 & 19.20 & 79.07 & 19.80 \\
gin\_ppo\_step & 23.20 & 49.30 & 27.53 & 18.87 & 72.87 & 19.60 \\
gin\_reinforce\_step & 23.25 & 49.30 & 27.47 & 18.87 & 72.80 & 19.60 \\
gin\_sac\_step & 23.05 & 50.50 & 27.60 & 20.13 & 74.00 & 21.70 \\
gin\_sarsa\_step & 23.20 & 50.10 & 27.60 & 19.07 & 73.60 & 19.90 \\
gat\_a2c\_step & 23.20 & 49.10 & 27.47 & 18.87 & 72.80 & 19.60 \\
gat\_dqn\_step & 24.50 & 51.00 & 31.80 & 20.00 & 92.60 & 21.20 \\
gat\_ppo\_step & 21.80 & 52.80 & 27.93 & 19.40 & 76.60 & 21.00 \\
gat\_reinforce\_step & 23.20 & 49.10 & 27.47 & 18.87 & 72.80 & 19.60 \\
gat\_sac\_step & 23.95 & 51.70 & 28.53 & 20.27 & 77.33 & 22.30 \\
gat\_sarsa\_step & 24.10 & 51.20 & 32.40 & 20.33 & 95.00 & 22.50 \\
gcn\_a2c\_step & 23.20 & 49.10 & 27.47 & 18.87 & 72.80 & 19.60 \\
gcn\_dqn\_step & 23.85 & 55.00 & 30.40 & 20.53 & 97.93 & 23.40 \\
gcn\_ppo\_step & 23.20 & 49.40 & 27.53 & 18.87 & 73.53 & 19.60 \\
gcn\_reinforce\_step & 24.05 & 52.50 & 27.87 & 20.13 & 84.13 & 22.70 \\
gcn\_sac\_step & 23.20 & 49.10 & 27.47 & 18.87 & 72.80 & 19.60 \\
gcn\_sarsa\_step & 23.20 & 49.10 & 28.33 & 18.87 & 73.00 & 19.60 \\
gin\_a2c\_step & 23.20 & 49.10 & 27.47 & 18.87 & 72.87 & 19.60 \\
gin\_dqn\_step & 24.15 & 66.10 & 31.87 & 20.33 & 97.20 & 21.10 \\
gin\_ppo\_step & 23.20 & 49.10 & 27.47 & 18.87 & 72.87 & 19.60 \\
gin\_reinforce\_step & 23.20 & 49.10 & 27.47 & 18.87 & 72.80 & 19.60 \\
gin\_sac\_step & 23.80 & 50.40 & 27.73 & 20.40 & 73.60 & 21.80 \\
gin\_sarsa\_step & 24.15 & 53.00 & 30.40 & 20.07 & 76.53 & 20.80 \\
v3\_generalist & 23.20 & 49.10 & 27.47 & 18.87 & 72.80 & 19.60 \\
\midrule
FFD & 23.20 & 49.10 & 27.47 & 18.87 & 72.80 & 19.60 \\
OPT & 20.00 & 48.30 & 27.33 & 17.33 & 72.07 & 18.70 \\
\bottomrule
\end{tabular}}\end{table}

\begin{table}[htbp]\centering
\caption{Stochastic beam ($B{=}20$, three seeds, mean$\pm$std average bins) for the policy models on the two discriminating families, sorted by Falkenauer\_T. FFD and GGA references shown at the bottom.}
\label{tab:fullbeam}\scriptsize
\begin{tabular}{lcc}
\toprule
Model & Falkenauer\_T & Scholl\_2 \\
\midrule
v2-gin/a2c & $21.13{\pm}0.08$ & $17.64{\pm}0.03$ \\
v2-gin/reinforce & $21.13{\pm}0.02$ & $17.69{\pm}0.08$ \\
v1-gin/ppo & $21.15{\pm}0.04$ & $17.58{\pm}0.03$ \\
v1-gin/reinforce & $21.25{\pm}0.04$ & $17.60{\pm}0.00$ \\
v2-gcn/a2c & $21.32{\pm}0.05$ & $17.96{\pm}0.03$ \\
v3-gcn/ppo & $21.35{\pm}0.11$ & $17.67{\pm}0.05$ \\
v2-gcn/sac & $21.43{\pm}0.08$ & $17.91{\pm}0.03$ \\
v2-gin/ppo & $21.48{\pm}0.06$ & $17.62{\pm}0.08$ \\
v2-gat/reinforce & $21.53{\pm}0.06$ & $17.73{\pm}0.05$ \\
v1-gat/reinforce & $21.58{\pm}0.06$ & $18.02{\pm}0.03$ \\
v1-gat/ppo & $21.63{\pm}0.06$ & $17.73{\pm}0.05$ \\
v1-gcn/sac & $21.72{\pm}0.09$ & $18.31{\pm}0.08$ \\
v2-gat/ppo & $21.77{\pm}0.02$ & $19.16{\pm}0.03$ \\
v1-gcn/reinforce & $21.83{\pm}0.16$ & $18.20{\pm}0.00$ \\
v1-gin/sac & $21.83{\pm}0.08$ & $18.00{\pm}0.14$ \\
v1-gcn/a2c & $21.87{\pm}0.06$ & $17.82{\pm}0.03$ \\
v1-gat/a2c & $21.93{\pm}0.10$ & $18.02{\pm}0.06$ \\
v2-gcn/ppo & $22.40{\pm}0.15$ & $18.49{\pm}0.03$ \\
v2-gat/sac & $22.53{\pm}0.02$ & $18.49{\pm}0.03$ \\
v2-gat/a2c & $22.70{\pm}0.04$ & $18.49{\pm}0.03$ \\
v1-gat/sac & $22.85{\pm}0.07$ & $18.62{\pm}0.08$ \\
v1-gcn/ppo & $22.85{\pm}0.04$ & $18.29{\pm}0.08$ \\
v2-gin/sac & $23.40{\pm}0.07$ & $18.87{\pm}0.11$ \\
v1-gin/a2c & $23.58{\pm}0.05$ & $18.87{\pm}0.05$ \\
v2-gcn/reinforce & $23.88{\pm}0.06$ & $19.36{\pm}0.03$ \\
\midrule
GGA (ref.) & 21.00 & 17.33 \\
FFD & 23.20 & 18.87 \\
\bottomrule
\end{tabular}\end{table}

\clearpage  %

\bibliographystyle{elsarticle-num}
\bibliography{references}

\end{document}